\documentclass[runningheads]{llncs}
\usepackage[T1]{fontenc}
\usepackage{graphicx}
\usepackage{amsmath}
\usepackage{multirow}
\usepackage{tabularx}
\usepackage{array}
\usepackage{xurl}
\usepackage[colorlinks=true,urlcolor=blue]{hyperref}
\usepackage{array}
\newcommand{\var}[1]{\texttt{\url{#1}}} 

\usepackage{orcidlink}
\renewcommand{\orcidID}[1]{\orcidlink{#1}}
\hypersetup{
    colorlinks=true,
    linkcolor=black,
    urlcolor=black,
    citecolor=black
}

\renewcommand{\arraystretch}{0.9}

\begin{document}

\title{Evolutionary Rule Extraction from Corporate Default Prediction Models}

\titlerunning{Evolutionary Rule Extraction from Corporate Default Prediction Models}

\author{Desiré Fabbretti\inst{1,2}\orcidID{0009-0007-8308-3145} \and
Matteo Pasquino\inst{1}\orcidID{0009-0005-5696-3214} 
Elia Pacioni\inst{2,3}\orcidID{0000-0002-1557-4870}
Caterina Lucarelli\inst{1}\orcidID{0000-0002-2632-878X} \and
Davide Calvaresi\inst{2}\orcidID{0000-0001-9816-7439}
}
\authorrunning{D. Fabbretti et al.}
%
\institute{Department of Management, Università Politecnica delle Marche, Ancona, Italy \\
\email{S1116451@studenti.univpm.it, m.pasquino@pm.univpm.it, c.lucarelli@staff.univpm.it} \and
HES-SO Valais-Wallis, 3960 Sierre, Switzerland \\
\email{\{elia.pacioni, davide.calvaresi\}@hevs.ch} \and
University of Extremadura, 06800 Mérida, Spain}
\maketitle              
\begin{abstract}
Small and medium-sized enterprises (SMEs) represent the majority of firms in most economies and often face financial constraints and higher vulnerability to financial distress. Predicting SME default is therefore crucial for financial institutions, policymakers, and researchers. Recent advances in machine learning (ML) have improved predictive performance in credit risk modeling. Yet, the limited interpretability of complex models raises concerns regarding transparency and regulatory compliance.
This study investigates SME's default predictors and applies explainable artificial intelligence (XAI) techniques to them. Using a panel of 50,718 Italian SME over the period 2015–2024, we compare traditional econometric approaches with several ML classifiers. The empirical results show that ML models significantly outperform the traditional logistic regression benchmark in terms of Balanced Accuracy and PR-AUC.
To address the interpretability challenge, we introduce DEXiRE-EVO, a novel evolutionary rule extraction framework that combines multi-objective optimization with the Contextual Importance and Utility (CIU) explainability method. The extracted rules reveal economically meaningful patterns associated with SME financial distress, highlighting the roles of weak internal liquidity generation, internal capital erosion, high leverage, and operational inefficiency. Additionally, contextual macroeconomic conditions and the persistence of financial instability contribute to identifying high-risk firms.
In general, the results show that combining ML with evolutionary rule extraction can improve both predictive performance and interpretability in credit risk modeling, thus supporting more transparent, data-driven decision-making in financial environments.

\end{abstract}
%
%
%

\section{Introduction}
\label{ch:intro}

Small and Medium Enterprises (SME) are an economic driving force, generating employment and contributing to value creation~\cite{Crosato2023}. Due to their limited access to alternative sources of finance, SME rely deeply on external financing (predominantly bank-intermediated). Therefore, identifying the SME's default determinants is impelling for researchers, financial institutions, and policymakers~\cite{cheraghali2024}.

Researchers have broadly investigated firms' default and financial distress, highlighting the importance of firm-level characteristics and aggregate economic conditions~\cite{Ciampi2021}. Traditional approaches leveraging accounting information and econometric models continue to provide a well-established framework for benchmarking and default analysis~\cite{Bazzana2024}.
Nonetheless, the expanding availability of firm-level data, combined with advances in computational techniques (i.e., ML), has sparked growing interest in more flexible, data-driven approaches to default prediction. Such approaches have shown potential improvements in predictive performance and credit risk analysis~\cite{Bazzana2024,Crosato2023,Nguyen2025,Zedda2024}. However, to date, the limited empirical evidence leaves their contribution debatable (exacerbated by firm heterogeneity, data constraints, and context-specific factors). 

As a consequence, many empirical studies still rely on limited or cross-sectional samples that may not fully capture the heterogeneity and temporal dynamics characterizing SME populations. Resampling strategies are often used to address class imbalance, but they may introduce biases that affect model generalizability. 
Although XAI techniques have increasingly been applied to credit risk models~\cite{Chang2025,Crosato2023}, most existing approaches focus on feature attribution rather than on reconstructing the model's underlying decision logic to understand how complex predictive models generate their decisions. Moreover, existing rule-extraction and evolutionary explainability methods are often not fully model-agnostic and have been evaluated solely in synthetic or low-dimensional settings, leaving open questions about their applicability to high-dimensional, imbalanced financial data. As predictive models become increasingly complex, the trade-off between accuracy and interpretability has emerged as a crucial challenge. This challenge is especially relevant for SME credit assessment, where transparency and economic plausibility are essential for sound risk management and regulatory compliance. To address this uncertainty, approaches that balance predictive performance with interpretable representations of the decision logic underpinning model behavior are required. 

To this end, this study proposes DEXiRE-EVO, a novel framework leveraging evolutionary computation to extract rules from opaque predictive models for SME default prediction. In particular, it presents (C1) a large-scale empirical evaluation of SME default prediction in a realistic, high-dimensional setting characterized by strong class imbalance and informational heterogeneity; and (C2) a novel post-hoc explainability approach based on evolutionary rule extraction guided by the CIU~\cite{anjomshoae2020pyciu} framework, enabling the reconstruction of transparent decision rules that preserve both predictive fidelity and economic interpretability.

The remainder of the paper is structured as follows.
Section~\ref{sec:sota} overviews the relevant literature. Section~\ref{sec:meth} describes the methodology. Section~\ref{sec:res} presents the empirical results. Section~\ref{sec:conclusion} concludes the paper.

\section{Related work}
\label{sec:sota}

This section reviews the main literature on SME default prediction. It first elaborates on traditional parametric models used in credit risk analysis, then examines the growing use of ML approaches, and finally considers recent developments in XAI and rule-extraction methods to improve the transparency of complex predictive models.

Modeling default risk has traditionally relied on parametric approaches, such as discriminant analysis and logistic regression~\cite{Altman1968,Ohlson1980}, which are primarily based on financial ratios and accounting variables. 
For decades, these methodologies have served as the benchmark for empirical default risk analysis~\cite{Ciampi2021,Crosato2023}. However, recent literature emphasizes that such models may be limited in their ability to capture complex, nonlinear relationships among explanatory variables, particularly in environments characterized by increasing informational heterogeneity and structural changes in credit markets~\cite{Moscatelli2020}.

Against this backdrop, ML methodologies have gained growing prominence (especially in the context of SME) where informational opacity and the relevance of qualitative relational factors complicate the modeling of insolvency risk~\cite{Brighi2018}.
The most adopted algorithms in credit risk modeling include ensemble methods, particularly Random Forest and Gradient Boosting, and Artificial Neural Networks. Random Forest~\cite{Breiman2001} is widely used, owing to its ability to model nonlinearities and higher-order interactions among multiple decision trees on random data subsamples. Boosting methods~\cite{Freund1997}, later extended through gradient-based formulations~\cite{Friedman2001}, construct sequential additive models that iteratively correct previous errors. XGBoost is a computationally efficient, regularized implementation of gradient boosting~\cite{Chen2016}. Artificial Neural Networks approximate nonlinear relationships through multilayer architectures and are particularly suitable for large and complex datasets~\cite{Rosenblatt1958,Rumelhart1986}.

The application of these techniques to corporate default prediction, especially for SME, has expanded significantly in recent years, frequently yielding improvements over traditional parametric specifications. Studies such as Crosato et al.~\cite{Crosato2023} and Zedda~\cite{Zedda2024} show that boosting algorithms, such as XGBoost, can outperform logit and probit models in terms of discriminatory power because they can model nonlinearities and higher-order interactions among financial, legal, and macroeconomic variables. Similarly, Moscatelli et al.~\cite{Moscatelli2020} report superior predictive performance of Random Forest and Gradient Boosting, particularly in settings with limited or less-structured information. In the context of Italian small- to medium-sized firms, Bazzana et al.~\cite {Bazzana2024} compare several ML classifiers with logistic regression, documenting a moderate improvement in predictive performance for ML models relative to the traditional specification. Consistent evidence is provided by Bitetto et al.~\cite{Bitetto2023}, who report the superior performance of a Random Forest–based approach compared with a parametric model in SME credit risk assessment. Moreover, Gu et al.~\cite{Gu2024} show that integrating resampling techniques with boosting algorithms enables more effective handling of class imbalance -- a structural feature of default prediction -- thereby improving the model's ability to identify distressed firms correctly.

Despite these documented gains in predictive performance, the increasing reliance on ML raises critical concerns regarding model transparency and interpretability. Crosato et al.~\cite{Crosato2023} argue that much of the recent literature has focused primarily on improving classification accuracy relative to linear benchmarks, leaving interpretability this context, the XAI literature distinguishes between interpretability, understood as intrinsic model transparency, and explainability, referring to post-hoc techniques designed to provide insight into models characterized by greater structural complexity~\cite{DoshiVelez2017,Lipton2018,Montavon2018}.
This distinction is particularly relevant in regulated financial environments, where transparency in the decision-making process is a substantive requirement, as emphasized by the European Commission's Ethics Guidelines for Trustworthy AI~\cite{EuropeanCommission2019}.

Explainability has been widely recognized as a crucial requirement in credit risk management. Scholars argue that XAI tools can help reduce information asymmetries between SME and potential lenders; consequently, explainability becomes essential for ensuring that ML applications in financial services and credit risk remain transparent and regulatory-compliant~\cite{Bitetto2024}. Building on these perspectives, Chang et al.~\cite{Chang2025} examine explainability tools for nonlinear credit scoring models and show that the quality and stability of explanations constitute an additional dimension beyond predictive accuracy. In particular, they demonstrate that when classes are strongly imbalanced (as is typical in default prediction), the stability of explanation methods may deteriorate, affecting both variable ranking and estimated feature contributions. These findings shift the focus from pure discriminative performance to the structural consistency and robustness of the model's learned decision logic.

Model-agnostic explainability methods, such as SHAP~\cite{lundberg2017unified} and LIME~\cite{ribeiro2016should}, have become widely adopted tools for improving the transparency of ML models. By quantifying feature contributions at both local and global levels, these methods aim to preserve predictive performance while increasing transparency. However, merely applying interpretative tools does not fully resolve the explainability challenge. As shown by Chang et al.~\cite{Chang2025}, explanation stability may be sensitive to data characteristics, particularly class imbalance, suggesting that credit risk model assessment should incorporate not only predictive accuracy but also the coherence and robustness of the explanations provided.

Recent advances include ECLAIRE~\cite{zarlenga2021efficientdecompositionalruleextraction}, which induces layer-wise rules via clause-wise substitution for deep neural networks, achieving high fidelity while sometimes producing longer rule sets. DEXiRE extends this paradigm by binarizing hidden layers to induce Boolean functions, producing concise propositional rules with improved execution efficiency while maintaining comparable accuracy and fidelity~\cite{electronics11244171}. Complementarily, other feature attribution approaches grounded in cooperative game theory provide theoretically principled explanations with desirable properties such as local accuracy and consistency~\cite{lundberg2017unified}. CIU offers an alternative model-agnostic framework that produces stable, context-aware explanations by quantifying both feature importance and utility relative to the prediction context~\cite{anjomshoae2020pyciu}. 
Decision trees remain strong interpretable baselines, with recent work exploring data-driven methods for learning splitting criteria that improve the accuracy–interpretability trade-off in tree models~\cite{balcan2024learning}.
Evolutionary post-hoc methods, such as G-REX (Genetic Rule Extraction, applied to models including neural networks and SVMs for applications including credit scoring)~\cite{martens2007comprehensible} further enable scalable genetic optimization of interpretable rule sets.

Although recent literature has extensively compared traditional parametric models and ML algorithms for default prediction, including interpretability considerations, most studies concentrate on specific model architectures or isolated evaluation criteria. Systematic comparisons across heterogeneous models in highly imbalanced, high-dimensional information settings remain relatively scarce. The present study contributes to this literature by providing a comprehensive comparative analysis based on a panel of 50,718 Italian SME over the period 2015-2024, including 6,695 defaulted firms and 44,023 active firms, thus constituting a strongly imbalanced classification problem. Unlike studies that rely on restrictive ex post sampling strategies, our dataset includes the full population of firms meeting the European SME definition criteria, thereby preserving substantial heterogeneity in size, sector, and geographic distribution.

Overall, the existing literature highlights the growing predictive potential of ML for SME default prediction, while also emphasizing the challenges of model transparency and interpretability. However, the application of XAI techniques in credit risk modeling remains relatively limited and still evolving. In particular, relatively little attention has been paid to approaches that can reconstruct explicit decision rules from complex predictive models in realistic financial environments characterized by high dimensionality, heterogeneous firm information, and severe class imbalance. This gap motivates the need for empirical frameworks that jointly evaluate predictive performance and the interpretability of the decision logic underlying ML models.
\vspace{-10pt}
\section{Methodology}
\label{sec:meth}
\begin{figure}
    \centering
    \includegraphics[width=1\linewidth]{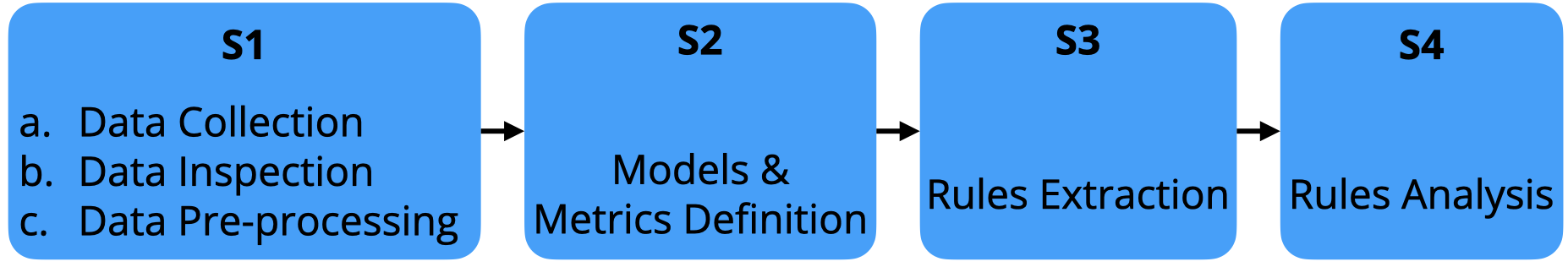}
    \caption{Organization of the methodological framework.}
    \label{fig:methodology}
\vspace{-0.5cm}
\end{figure}

\vspace{-10pt}
This section presents the empirical methodology adopted in the study. The methodological framework is organized into four sections (S1–S4), see Figure~\ref{fig:methodology}.
\vspace{-20pt}

\noindent \textbf{S1a,b: Data collection \& inspection:} The empirical analysis relies on firm-level data drawn from AIDA (Bureau van Dijk ---\href{https://login.bvdinfo.com/R1/AidaNeo}{link}), which provides accounting and legal information based on Italian firms' statutory financial statements. The unit of observation is the firm-year, and the dataset is organized as an unbalanced panel covering the period 2015-2024, subject to data availability.

The sample comprises $50,718$ Italian SME, identified using the quantitative thresholds established by the European Commission~\cite{eu2003} and applied during data extraction: firms with fewer than 250 employees and an annual turnover not exceeding EUR 50 million and/or total assets not exceeding EUR 43 million. The sample is further restricted by excluding firms operating in financial and insurance activities, real estate activities, public administration, education, human health, and social work activities, as well as extraterritorial organizations. Sectoral information is drawn from the ATECO 2007 classification, which corresponds to the Italian implementation of NACE Rev. 2, as reported by AIDA.

For each firm, the dataset includes identification and structural characteristics, including legal form, year of incorporation, geographic region, and primary sector of activity. Accounting information is drawn from firms' statutory financial statements and includes balance-sheet and income-statement data. These data provide information on firms' size, capital structure, liquidity, profitability, and operating performance. When available, accounting information is observed for up to ten consecutive fiscal years over the sample period.

In addition to firm-level variables, the empirical specification incorporates macroeconomic and contextual variables at the sectoral and regional levels, as previous research has emphasized the relevance of macro-level factors in default prediction~\cite{Zhou2023}. Specifically, the dataset includes sector-level default rates and value-added dynamics, regional gross domestic product (GDP) growth, harmonized bank lending rates on new loans to non-financial corporations, and consumer price inflation. Sectoral and regional indicators are drawn from official ISTAT statistics (\href{https://www.istat.it}{link}). At the same time, lending rates and sector-level default rates are based on harmonized series produced within the Eurosystem and distributed by the Bank of Italy (\href{https://www.bancaditalia.it}{link}). The model also accounts for structural firm characteristics treated as categorical variables -- namely, legal form, region, and sector classification -- and includes a procedural status indicator capturing whether the firm was under a court-supervised composition with creditors in year $t-1$. This indicator is used as a predictor of default in year t. Table~\ref{macro_variables} summarizes the macroeconomic and contextual variables included in the model.

\begin{table}[ht]
\vspace{-0.4cm}
\centering
\footnotesize
\caption{Macroeconomic and Contextual Variables included in the dataset.}
\label{macro_variables}
\begin{tabular}{ll}
\hline
\textbf{\begin{tabular}[c]{@{}l@{}}Macroeconomic \\ and Sectoral Controls\end{tabular}} & \begin{tabular}[c]{@{}l@{}}Inflation Rate, Regional GDP Growth Rate, \\ Sector Value Added Growth, Sector Default Rate\end{tabular} \\ \hline
\textbf{\begin{tabular}[c]{@{}l@{}}Firm Structural \\ Characteristics\end{tabular}} & Region, Legal Form, ATECO Code \\ \hline
\textbf{Firm Dimension Proxy} & Age, Ln Total Assets, Ln Sales \\ \hline
\textbf{Procedural Status} & Court-supervised Composition with Creditors Indicator \\ \hline
\end{tabular}
\end{table}
\vspace{-0.5cm}

The final sample includes 6,695 firms classified as default according to their legal status in AIDA, each with at least one accounting observation during the sample period, and 44,023 firms classified as active firms. Overall, the dataset contains 507,180 firm-year observations over the sample period.\\

\noindent \textbf{S1c: Data preprocessing:} After constructing the raw dataset, we implement a structured processing pipeline to obtain a firm-year panel suitable for default prediction while ensuring temporal consistency and preventing information leakage. The original wide-format accounting data are reshaped into a longitudinal structure, in which each observation corresponds to a uniquely identified firm-year pair. 
Following temporal filtering and data cleaning procedures, the final modeling sample comprises 472,242 firm-year observations, compared to 507,180 observations in the raw dataset. 

The sample is partitioned at the firm level into training (70\%), validation (20\%), and test (10\%) subsets using stratified sampling based on firm-level default status. This procedure ensures that all observations of a given firm belong to the same subset while preserving class proportions. Each split spans the full 2015-2024 period, so that all subsets are exposed to the same macroeconomic regime and recent shocks. The firm-level stratification preserves class proportions across splits, with nearly identical default rates in training (1.4173\%), validation (1.4179\%), and test (1.4203\%), closely matching the overall rate of 1.4177\%.

The target default is a binary indicator equal to one for firms classified as bankrupt or dissolved due to bankruptcy in AIDA. For firms defaulting in year $t$, the indicator is assigned to year $t-1$, so that information observed at time $t-1$ is used to predict the occurrence of default in year $t$. For defaulted firms, all observations with a year greater than or equal to the recorded default year $(t)$ are removed to avoid post-event contamination. 
All covariates, including financial variables, sectoral and regional indicators, and macroeconomic controls, are temporally aligned using information available up to $t-1$. Only pre-default information is retained for model estimation. This one-year-ahead setup enforces temporal consistency and precludes look-ahead bias.

A ratio-based feature specification is adopted, relying on financial indicators commonly used in the default prediction literature, including leverage, liquidity, profitability, working capital, turnover, and interest coverage measures.
Given their pronounced skewness and the presence of near-zero denominators that would otherwise generate undefined values, the interest coverage ratios (EBITDA/Interest Expenses and EBIT/Interest Expenses) are capped at the $99.5^{th}$ percentile of the training set (with the same cap subsequently applied to validation and test data) and transformed using the function $\operatorname{sign}(x)\,\log(1 + |x|)$, thereby preserving both sign and economic interpretability.

Table~\ref{ratio} reports the financial ratios for each firm in the sample and their definitions, providing a transparent overview of the variables used in the empirical specification.
 
\begin{table}[ht]
\centering
\scriptsize
\caption{Description of the financial indicators included in the dataset.}
\label{ratio}
\begin{tabular}{@{}l p{7cm}@{}}
\hline
\textbf{Index} & \textbf{Definition (Measured)} \\
\hline
Current Ratio & Short-term liquidity by comparing realizable assets to short-term obligations. \\
Net Working Capital to Total Assets & Net short-term liquidity relative to firm size. \\
Leverage & Proportion of assets financed through debt. \\
Debt-to-Equity & Debt financing's relative weight w.r.t. to equity. \\
Tax and Social Security Debt Ratio & Public-law obligations share within total liabilities. \\
Retained Earnings-to-Total Assets & Cumulative profitability \& internal capital accumulation. \\
EBITDA-to-Interest Expense & Operating cash-flow coverage of financial charges. \\
EBIT-to-Interest Expense & Earnings-based coverage of interest obligations. \\
Interest Expense-to-Total Debt & Effective cost of debt financing. \\
Return on Equity (ROE) & Profitability relative to equity capital. \\
Return on Assets (ROA) & Overall profitability relative to the asset base. \\
Cash Flow-to-Total Assets & Internally generated liquidity relative to firm size. \\
Turnover Ratio & Asset utilization efficiency. \\
\hline
\end{tabular}
\vspace{-0.6cm}
\end{table}

Beyond the baseline specification, two alternative dataset variants are constructed to investigate the relevance of macroeconomic and contextual variables within the XGBoost framework.
The first variant excludes macroeconomic and sector-level controls, allowing us to assess whether firm-level financial ratios primarily drive predictive performance and feature relevance or whether macro-contextual information materially contributes to the model's decision structure.

The second variant augments the baseline feature space with firm-specific exponentially weighted moving averages and standard deviations, computed in chronological order with a two-year half-life, thereby incorporating persistence and volatility dynamics.
These configurations are introduced to enable a structured evaluation of how macroeconomic variables and dynamic information affect feature importance patterns and decision logic in the subsequent analyses.

All preprocessing is embedded within model-specific training pipelines, estimated exclusively on the training set, and subsequently applied to validation and test data. 
Numerical variables are imputed using median imputation, while categorical data are encoded using ordinal encoding.\\

\noindent \textbf{S2: Models and metrics definition} We estimate four classifiers: Logistic Regression, Random Forest, XGBoost, and Multi-Layer Perceptron (MLP) under the common train/validation/test partition. 
Hyperparameter configurations are selected based on validation-set performance, as measured by Balanced Accuracy and the Precision–Recall Area Under the Curve (PR-AUC). 
Classification thresholds are calibrated separately on the validation set. Final performance is evaluated exclusively on the held-out test set to ensure an unbiased out-of-sample assessment. The importance of systematic hyperparameter optimization relative to default configurations is well established in the ML literature~\cite{Bergstra2012}.
Given the rarity of default events, both hyperparameter selection and threshold calibration explicitly address class imbalance. Model performance is primarily evaluated using Balanced Accuracy, defined as the arithmetic mean of sensitivity and specificity. This metric is particularly appropriate in imbalanced settings, as it assigns equal weight to correctly classifying default and non-default firms. In contrast, standard accuracy can be misleading in rare-event contexts, as high values may reflect the model's ability to correctly classify the majority class (active firms) while failing to identify default events adequately. PR-AUC, being threshold-independent, provides a complementary evaluation of ranking performance in rare-event settings.

The conventional 0.5 decision threshold is not adopted, as it is generally inappropriate in the presence of marked class imbalance. Instead, the classification cutoff is selected on the validation set by maximizing the G-mean (the geometric mean of sensitivity and specificity), following the approach adopted by Bazzana et al.~\cite{Bazzana2024}. The selected threshold is then held fixed for out-of-sample evaluation on the test set.\\

\noindent \textbf{S3: Explainers \& rules extraction} 
DEXiRE was originally designed as a rule extraction pipeline for neural models, in which rule induction is performed using intermediate representations (layer activations) that serve as an explanatory latent space. To extend its applicability to tree-based models such as \texttt{XGBoost}, which lack an internal hierarchy of representations directly usable for interpretability purposes, the framework has been generalized to become model-agnostic. Through a unified adaptation interface, DEXiRE can automatically select the most appropriate extraction strategy, distinguishing between models exposing internal layers and those that do not. In the latter case, a direct extraction mode in the input feature space is adopted, where rules are induced from the pair \((X, \hat{y}_{\text{model}})\) to approximate the predictive behavior of the original model and produce rule-based explanations for non-neural algorithms as well.

This model-agnostic approach serves as conceptual inspiration for DEXiRE-EVO, which incorporates CIU-guided multi-objective evolutionary optimization to derive high-fidelity interpretable rule sets from black-box models. The objective is to transform the behavior of a black-box classifier into a symbolic representation composed of decision rules, while maintaining a controlled trade-off between predictive fidelity and interpretability.

The method operates in a post-hoc setting: the predictive model is treated as a fixed system, and DEXiRE-EVO learns a set of rules that approximates its decision behavior. To do so, the framework is organized into three main stages, each building upon the previous one.

\textbf{Contextual relevance estimation.}  
First, the contextual importance and utility of input variables are estimated using the CIU framework, which provides a semantic measure of feature relevance for each prediction by quantifying both a feature's importance within the prediction context and its contribution to the predicted outcome.

\textbf{Evolutionary search for rule sets.}  
Second, a population of candidate rule sets is generated and iteratively evolved using a multi-objective evolutionary optimization procedure based on the NSGA-II algorithm. Each individual represents an ordered set of decision rules approximating the behavior of the original model.

\textbf{Pareto-optimal rule selection.}  
At the final stage, non-dominated solutions are identified along the Pareto frontier. From these, a final rule set is selected that maximizes fidelity with respect to the black-box model while preserving interpretability.

The outcome of these stages is a set of explanations organized as an ordered decision list. Each rule consists of a conjunction of predicates over input variables:
\[
(x_1 \leq \tau_1 \land x_2 > \tau_2 \land \dots \land x_k \leq \tau_k) \rightarrow y = 1
\]

where $\tau_i$ represents a threshold value associated with variable $x_i$. The alternative class is assigned through a default rule (else $\rightarrow$ class 0).

The multi-objective formulation of DEXiRE-EVO optimizes two objectives simultaneously:

\begin{itemize}
\item Fidelity: the ability of the rule set to replicate the predictions of the black-box model, measured using balanced accuracy;
\item CIU alignment: the semantic consistency between the variables appearing in the rules and their contextual importance as estimated by CIU.
\end{itemize}

New candidate solutions are generated through evolutionary operators. The crossover operator is applied in two complementary forms:  
(i) exchange of segments of the decision list between two parent individuals, allowing different rule combinations to emerge;  
(ii) exchange of logical sub-expressions within rules, enabling recombination of local decision patterns.

The mutation operator modifies both the parameters and the structure of rules. In particular, mutation can: (i) modify predicate thresholds, (ii) change comparison operators, (iii) replace the variable used in a predicate, (iv) add or remove predicates, (v) add or remove rules from the decision list.

In the configuration adopted in this study, logical connectives between predicates are restricted to $\land$ (AND), and the positive class is fixed to $y=1 (DEFAULT)$. Conversely, the alternative class is assigned by the default rule.

The parameters of the evolutionary algorithm were determined via a grid search and are designed to balance convergence and exploration of the solution space. The evolutionary search is initialized with a population of 120 individuals and runs for up to 500 generations. The crossover probability is set to 0.7 and the mutation probability to 0.5. An early stopping mechanism is adopted when no improvement in the Pareto frontier is observed for 25 consecutive generations. Additionally, the algorithm preserves 5 elite solutions to maintain high-quality rule sets across generations.

Overall, DEXiRE-EVO performs interpretable distillation of the predictive model via a CIU-guided evolutionary search. The resulting rule sets provide transparent, human-readable explanations that approximate the original model's decision behavior while maintaining strong predictive fidelity. 

In the following experiments, DEXiRE and DEXiRE-EVO are compared primarily on predictive fidelity to the black-box model\\

\noindent \textbf{S4: Rule-sets Analysis} From the resulting rules generated by DEXiRE-EVO, a sample of 21 rules was extracted. Then, five representative rules were selected, based on predictive fidelity and CIU alignment values. Preference was given to rules involving the most informative variables identified in the feature contribution analysis. The selected rules were then interpreted from an economic perspective, linking their logical structure to the previously identified financial vulnerability patterns. Finally, the rules were jointly analyzed to reconstruct the model's overall reasoning structure and to assess its coherence with the feature-importance and dataset-variant comparison results.

\section{Results}
\label{sec:res}
This section is organized into four stages. First, we compare the predictive performance of the estimated models. Given its superior out-of-sample performance, XGBoost is selected for the explainability analysis, which investigates feature impact using CIU and derives the model's decision rules.

\subsection{Comparative Predictive Performance}
Predictive performance is assessed for Logistic Regression, Random Forest, XGBoost, and MLP under the baseline ratio-based specification. All metrics are computed on the held-out test set, with classification thresholds calibrated on the validation set to maximize the G-mean.

Given the rarity of corporate defaults in our sample (1.4\%), model performance is assessed using Balanced Accuracy and the Precision-Recall Area Under the Curve (PR-AUC). 
Balanced Accuracy provides a threshold-dependent measure that assigns equal weight to sensitivity and specificity
\[
\text{Balanced Accuracy} = \frac{\text{Sensitivity} + \text{Specificity}}{2}.
\]

Sensitivity measures the proportion of correctly identified defaulting firms (true positive rate), whereas specificity measures the proportion of correctly classified non-defaulting firms (true negative rate). 

PR-AUC evaluates the model's ability to rank firms according to default risk by summarizing the trade-off between precision and recall across all possible thresholds and, in rare-events contexts, provides a particularly informative measure because it focuses on performance with respect to the minority class without being inflated by the large number of correctly classified non-defaulting firms.
Its magnitude should be interpreted relative to the sample's default prevalence. With a default rate of approximately 1.4\%, a random classifier would achieve a value close to 0.014. Substantially higher values, therefore, indicate that the model effectively concentrates true default events among firms assigned higher predicted risk. Table~\ref{tab:performance} summarizes the out-of-sample performance of the four classifiers in terms of Balanced Accuracy and PR-AUC.

\begin{table}[ht]
\vspace{-0.3cm}
\centering
\caption{Out-of-sample predictive performance of the classification models considered in the empirical analysis. The table reports test-set results for Logistic Regression, Random Forest, XGBoost, and a Multilayer Perceptron (MLP) in terms of Balanced Accuracy and PR\_AUC.}
\label{tab:performance}
\small
\renewcommand{\arraystretch}{1.2}
\begin{tabular}{|c|c|c|c|c|}
\hline
 & Logit & Random Forest & XGBoost & MLP \\
\hline
Balanced Accuracy & 0.551 & 0.880 & 0.901 & 0.852 \\
\hline
PR\_AUC & 0.016 & 0.390 & 0.429 & 0.262 \\
\hline
\end{tabular}
\vspace{-0.6cm}
\end{table}

XGBoost achieves the highest Balanced Accuracy (0.901), indicating the most effective balance between correctly identifying defaulting firms and correctly classifying non-defaulting firms.
Random Forest follows closely (0.880), confirming the strong performance of tree-based ensemble methods in modeling default risk.
The MLP achieves a lower but still substantial value (0.852).
Logistic Regression exhibits markedly weaker performance (0.551), consistent with recent empirical evidence showing that nonlinear models outperform traditional linear classifiers in corporate default prediction.

A similar result emerges when performance is evaluated using PR-AUC. XGBoost again delivers the strongest result (0.429), confirming its superior ability to rank firms according to default risk in a highly imbalanced context.
Random Forest achieves a slightly lower value (0.390), while MLP records a more moderate ranking performance.
Logistic Regression performs only marginally better than the baseline prevalence of the default (0.016 versus approximately 0.014), indicating limited discriminative power for the minority class.

Taken together, these results show the consistent dominance of XGBoost in terms of both Balanced Accuracy and PR-AUC, supporting its selection as the reference model for the subsequent explainability analysis.

\subsection{Datasets and Selected Features' Impact}

While Section 4.1 compares predictive performance across model architectures, this section examines how the dataset's informational structure affects the behavior of the best-performing model, XGBoost. In particular, we assess whether predictive performance is driven primarily by firm-level financial ratios or by contextual and historical information that provides additional predictive signal. Although many studies rely primarily on firm-level accounting indicators, external economic conditions may also introduce systematic risk components not fully captured by balance-sheet variables.

To this end, we construct three alternative dataset configurations. The first corresponds to the baseline dataset (D1). The second configuration (D2) excludes macroeconomic variables, retaining only firm-level financial ratios and structural features. The third configuration (D3) extends the baseline specification by incorporating historical dynamics through exponentially weighted moving averages and standard deviations of financial ratios.

Model performance is evaluated using Balanced Accuracy and PR-AUC. Table~\ref{tab:D_performance} reveals two main patterns. First, macroeconomic and sectoral variables significantly improve predictive performance. The baseline dataset (D1) achieves a Balanced Accuracy of 0.901 compared with 0.879 for the specification without contextual information (D2). The difference is even more pronounced for PR-AUC, which declines from 0.429 to 0.366 when macroeconomic variables are removed. This suggests that contextual information provides a substantial additional signal for identifying financially vulnerable firms.

Second, incorporating historical summary statistics has little effect on Balanced Accuracy but improves ranking performance. The dataset including historical dynamics (D3) achieves a Balanced Accuracy of 0.903, very close to the baseline specification, while PR-AUC increases to 0.449. This indicates that historical financial trajectories refine the ranking of firms by default risk, although their impact on threshold-based classification remains limited.

\begin{table}[ht]
\centering
\caption{Performance comparison of the XGBoost model across the three dataset specifications: Standard (D1), without macroeconomic variables (D2), and with exponentially weighted mean and standard deviation features (D3).}
\label{tab:D_performance}
\begin{tabular}{llll}
\cline{2-4}
\multicolumn{1}{l|}{}                   & \multicolumn{1}{l|}{D1} & \multicolumn{1}{l|}{D2} & \multicolumn{1}{l|}{D3} \\ \cline{1-4}
\multicolumn{1}{|l|}{Balanced Accuracy} & \multicolumn{1}{l|}{0.901}   & \multicolumn{1}{l|}{0.879}   & \multicolumn{1}{l|}{0.903}        \\ \cline{1-4}
\multicolumn{1}{|l|}{PR-AUC}            & \multicolumn{1}{l|}{0.429}   & \multicolumn{1}{l|}{0.366}   & \multicolumn{1}{l|}{0.449}       \\ \cline{1-4}
\end{tabular}
\vspace{-0.7cm}
\end{table}

\begin{figure}[htbp]
    \centering
    
    \includegraphics[width=0.75\linewidth]{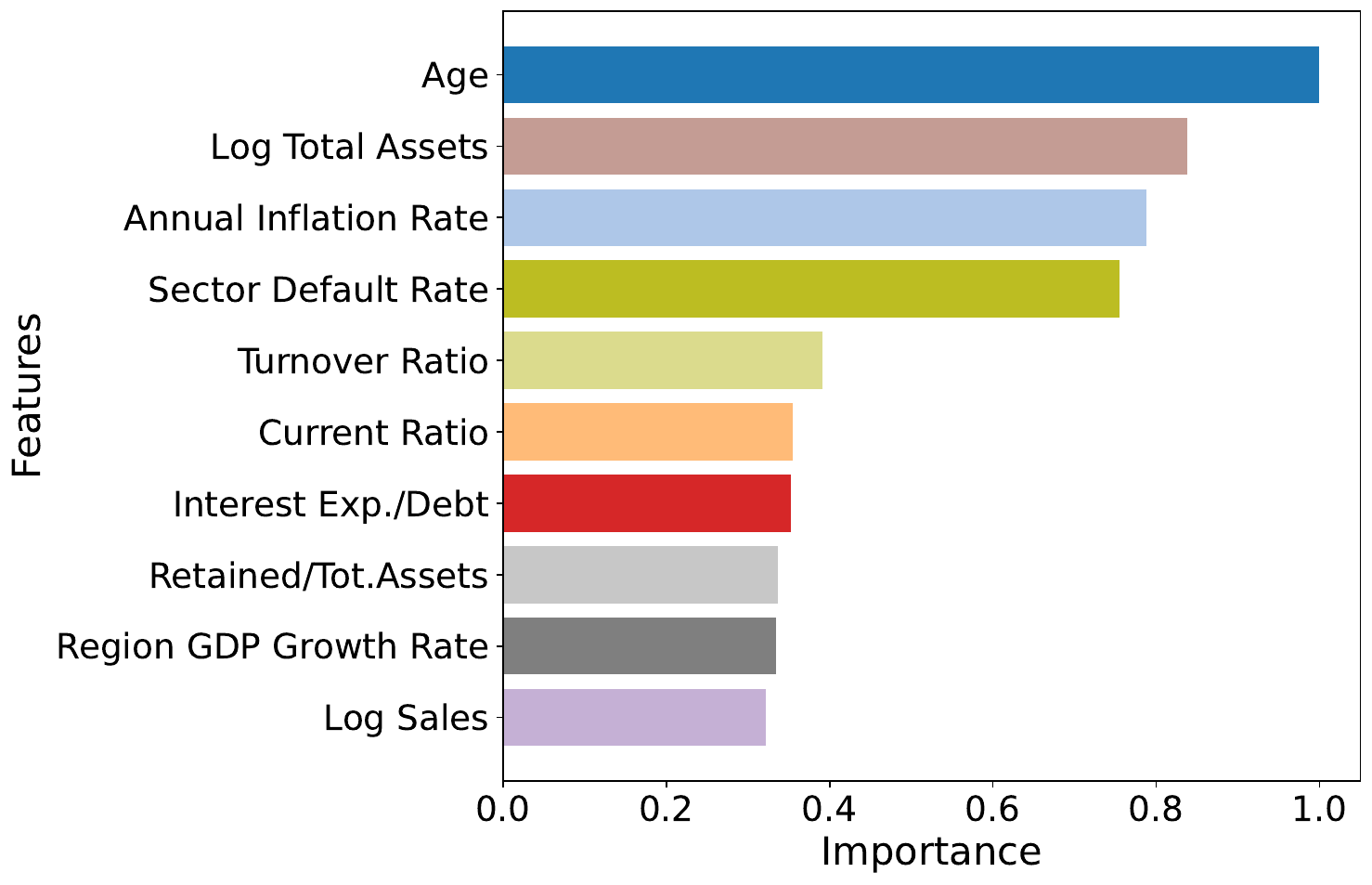}
    \caption{Feature importance ranking for dataset configuration D1.}
    \label{fig:D1_features}

    \vspace{0.4cm}

    \includegraphics[width=0.75\linewidth]{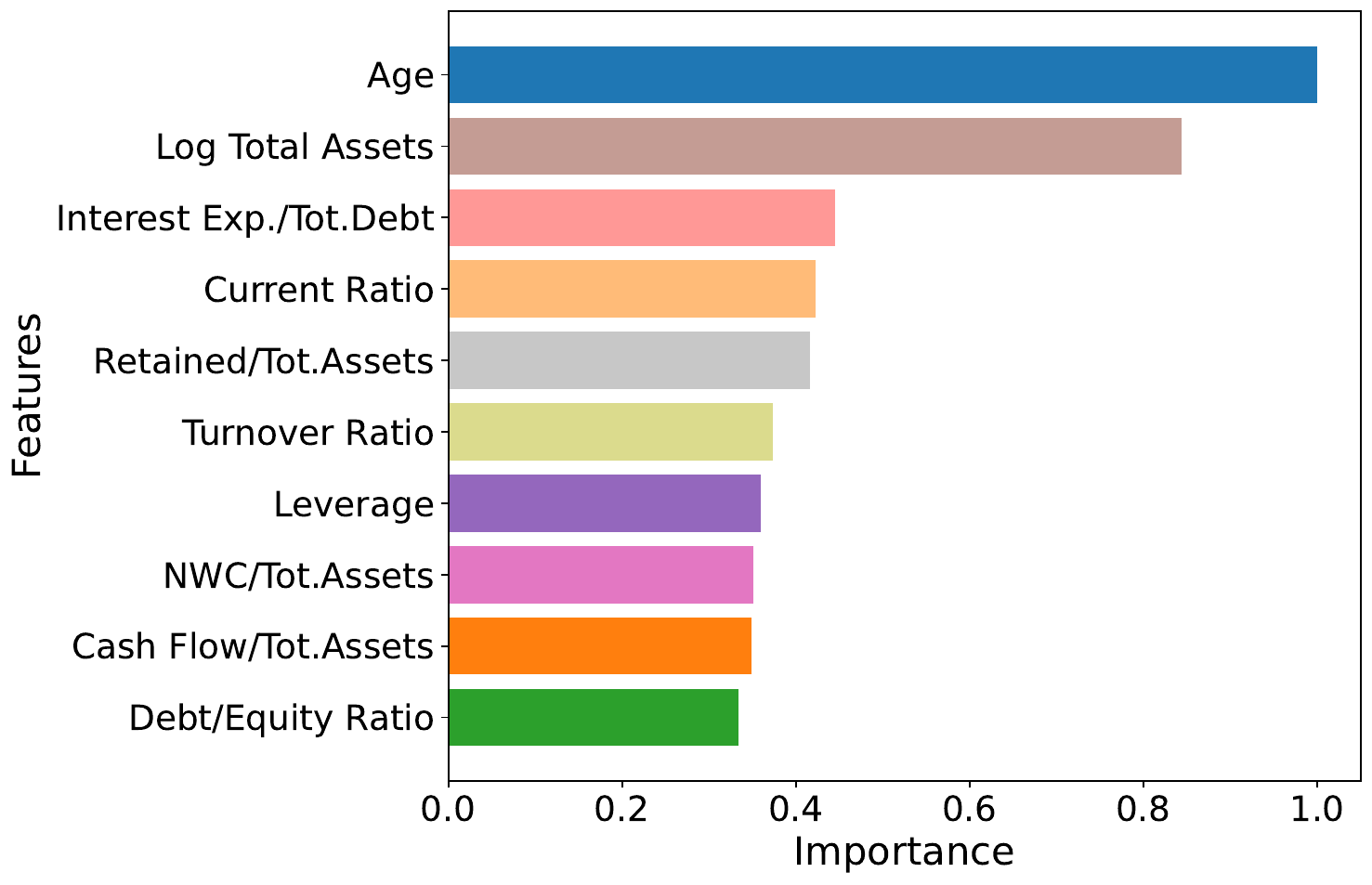}
    \caption{Feature importance ranking for dataset configuration D2.}
    \label{fig:D2_features}

    \vspace{0.4cm}

    \includegraphics[width=0.75\linewidth]{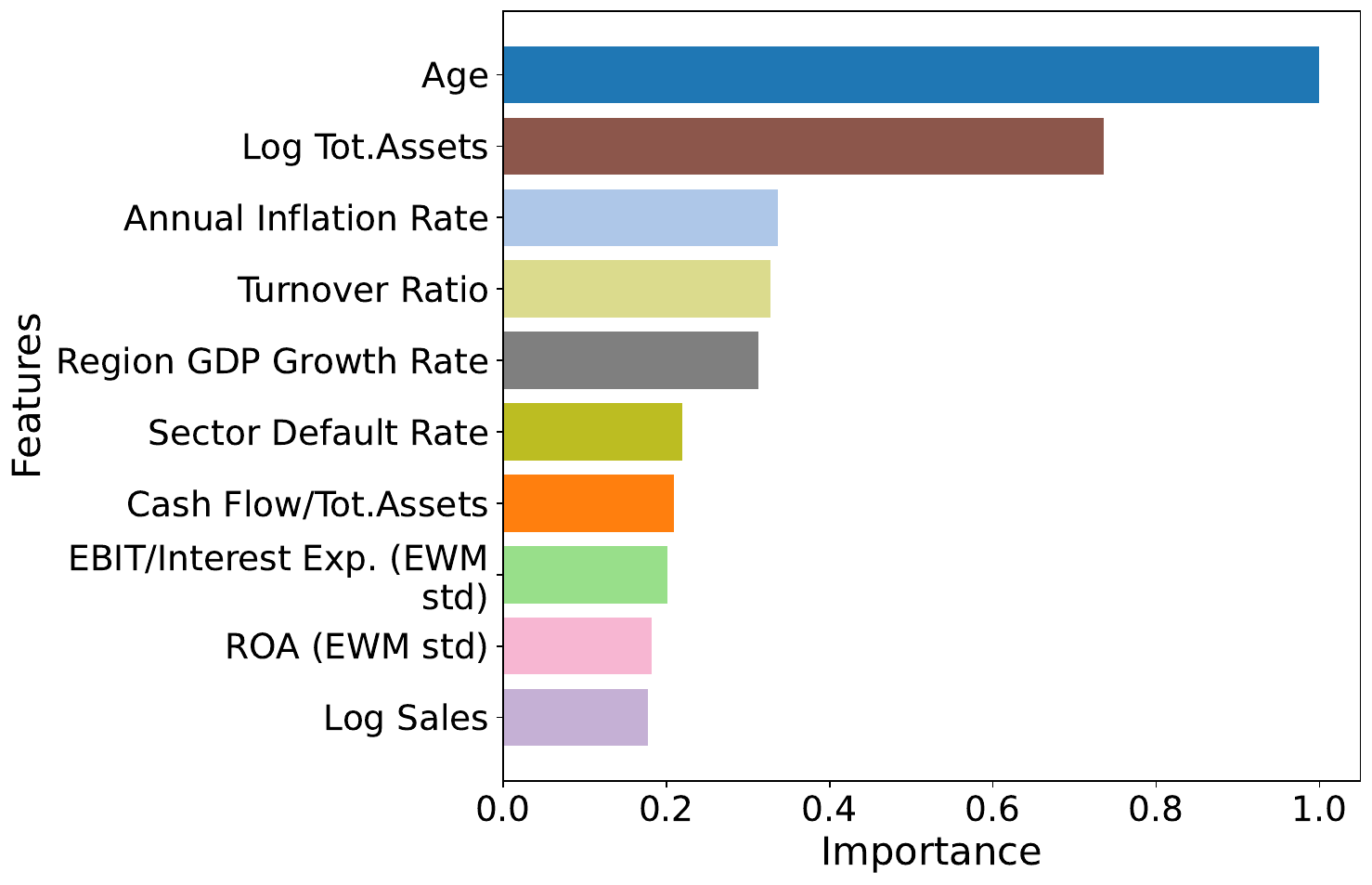}
    \caption{Feature importance ranking for dataset configuration D3.}
    \label{fig:D3_features}

\end{figure}

To further investigate how dataset structure affects the model's decision process, we conduct a feature contribution analysis using the CIU explainability framework. Figures~\ref{fig:D1_features}--\ref{fig:D2_features}--\ref{fig:D3_features} report the ranking of the ten most influential features across the three dataset configurations.

In the baseline configuration (D1), the most influential predictors are firm age and firm size (ln TA), which capture structural characteristics of SME and are widely recognized determinants of corporate survival. Several contextual variables also emerge among the top contributors, including annual inflation, sector default rate, and regional GDP variation, suggesting that the model captures systematic risk components affecting groups of firms simultaneously.

When macroeconomic variables are excluded (D2), the model relies exclusively on firm-level financial indicators, including cost of debt, liquidity measures (current ratio and net working capital), profitability proxies (retained earnings to total assets), and leverage indicators. Although the ranking of individual variables differs slightly from D1, the set of dominant financial predictors remains largely unchanged, indicating that the model relies on a similar financial information structure, whereas removing macroeconomic variables eliminates an additional contextual signal.

Finally, in the dataset that includes historical dynamics (D3), two historical variables — the standard deviation of EBIT-to-interest expenses and the standard deviation of ROA — are among the most influential predictors. These variables capture instability in firms' financial performance over time. Notably, volatility measures appear more influential than historical averages, suggesting that fluctuations in profitability and debt-servicing capacity represent particularly strong signals of financial distress.

Overall, the comparison of dataset configurations shows that the dataset's informational structure affects both predictive performance and the model's internal decision logic. Structural firm characteristics remain the dominant predictors across all configurations, whereas macroeconomic context and financial volatility provide complementary signals for identifying vulnerable firms.

\subsection{Explainers performance and selection}
The performance comparison between DEXiRE and DEXiRE-EVO focuses on predictive fidelity relative to the XGBoost classifier and CIU alignment. DEXiRE attains a fidelity of 0.832 and a CIU alignment of 0.444, whereas DEXiRE-EVO achieves higher fidelity (mean 0.856, std 0.004) and substantially improved CIU alignment (mean 0.684, std 0.096) across configurations (Table~\ref{tab:explainer_comparison}). This gain reflects the multi-objective design of DEXiRE-EVO, in which CIU alignment is explicitly incorporated into the evolutionary fitness function alongside fidelity.

\begin{table}[h]
\vspace{-0.3cm}
\centering

\caption{DEXiRE vs DEXiRE-EVO performance across XGBoost configurations.}
\label{tab:explainer_comparison}
\begin{tabular}{|l|c|c|c|c|}
\hline
\textbf{Method} & \textbf{Fidelity $\mu$} & \textbf{Fidelity $\sigma$} & \textbf{CIU $\mu$} & \textbf{CIU $\sigma$} \\
\hline
DEXiRE & 0.832 & -- & 0.444 & -- \\
DEXiRE-EVO (D1) & 0.854 & 0.002 & 0.697 & 0.071 \\
DEXiRE-EVO (D2) & 0.859 & 0.007 & 0.764 & 0.086 \\
DEXiRE-EVO (D3) & 0.854 & 0.003 & 0.591 & 0.084 \\
\textbf{DEXiRE-EVO (mean)} & \textbf{0.856} & \textbf{0.004} & \textbf{0.684} & \textbf{0.096} \\
\hline
\end{tabular}
\vspace{-0.4cm}
\end{table}

Based on these results, DEXiRE-EVO is adopted as the primary explainer for subsequent analysis. It provides rule sets that closely replicate XGBoost's decision behavior and remain better aligned with CIU-based feature relevance.

The application of DEXiRE-EVO to the XGBoost variants reveals distinct patterns in rule-extraction performance. The D2 configuration (no macro, fewer variables) yields the highest fidelity (0.859) and CIU alignment (0.764), confirming that simpler feature spaces facilitate more accurate rule extraction despite XGBoost's lower predictive performance on D2 (Table~\ref{tab:D_performance}). The D1 baseline exhibits robust explainability (fidelity 0.854, CIU 0.697), while the D3 EWM-augmented specification maintains comparable fidelity (0.854) despite greater feature complexity and improved XGBoost performance, albeit with reduced CIU alignment (0.591). These trade-offs highlight the value of historical dynamics in enhancing model expressiveness without severely compromising explainability, suggesting promising directions for temporal credit risk modeling.

\subsection{XGBoost Model's Decision Rules}
In addition to the CIU-based analysis presented in the previous section, we examine the decision rules extracted from the trained XGBoost model to interpret its predictions further. 
Representative rules are reported for the three dataset configurations introduced previously (D1, D2, and D3), illustrating how the model combines relevant variables under specific conditions and thresholds. The selected rules are presented in Table~\ref{tab:rules}, where each rule is accompanied by its corresponding fidelity and CIU alignment.

\begin{table}[htb!]
\centering
\tiny
\setlength{\tabcolsep}{3pt}
\renewcommand{\arraystretch}{1.05}
\caption{Representative rules extracted from the XGBoost model for the three dataset configurations (D1, D2, and D3), with corresponding fidelity and CIU alignment.}
\label{tab:rules}

\begin{tabularx}{\linewidth}{|c|c|c|c|>{\raggedright\arraybackslash}X|}
\hline
\textbf{ID} &
\textbf{\begin{tabular}[c]{@{}c@{}}Dataset\end{tabular}} &
\textbf{Fidelity} &
\textbf{CIU align} &
\textbf{Rules} \\
\hline

\textbf{1} &
\multirow[c]{2}{*}{\centering D1} &
0.853 & 0.76 &
$\begin{gathered}
(\mathrm{RETAINED}/\mathrm{TA} < -0.08 \land 9 \le \mathrm{AGE} < 41)\\[-3pt]
\lor\\[-3pt]
(\mathrm{AGE} \ge 9 \land \mathrm{CF\_TO\_ASSETS} \le 0.02)\\[1pt]
\end{gathered}$
\\ \cline{1-1}\cline{3-5}

\textbf{2} &
& 0.852 & 0.772 &
$\begin{gathered}
(\mathrm{ln\_TA} \ge 7.35 \land \mathrm{AGE} \ge 56 \land \mathrm{Sector\ Default\ Rate} \ge 0.03)\\[-3pt]
\lor\\[-3pt]
(9 \le \mathrm{AGE} < 56 \land \mathrm{NWC}/\mathrm{TA} \le -0.74)\\[-3pt]
\lor\\[-3pt]
(\mathrm{AGE} \ge 9 \land \mathrm{CF\_TO\_ASSETS} < 0.02)\\[1pt]
\end{gathered}$
\\ \hline

\textbf{3} &
\centering D2 &
0.867 & 0.757 &
$\begin{gathered}
(8 < \mathrm{AGE} \le 22 \land \mathrm{ln\_TA} \ge 6.50 \land \mathrm{RETAINED}/\mathrm{TA} < 0.00)\\[-3pt]
\lor\\[-3pt]
(\mathrm{AGE} > 8.00 \land \mathrm{CF\_TO\_ASSETS} < 0.01)\\[-3pt]
\lor\\[-3pt]
(\mathrm{AGE} \ge 9.00 \land \mathrm{DEBT}/\mathrm{EQUITY} > 23.71)\\[1pt]
\end{gathered}$
\\ \hline

\textbf{4} &
\multirow[c]{2}{*}{\centering D3} &
0.85 & 0.781 &
$\begin{gathered}
(\mathrm{RETAINED}/\mathrm{TA\_ewm\_mean} < -0.23 \land 9 \le \mathrm{AGE} < 32)\\[-3pt]
\lor\\[-3pt]
(\mathrm{CF\_TO\_ASSETS} < 0.01 \land \mathrm{AGE} \ge 9.00)\\[1pt]
\end{gathered}$
\\ \cline{1-1}\cline{3-5}

\textbf{5} &
& 0.857 & 0.624 &
$\begin{gathered}
(\mathrm{AGE} > 8.00 \land \mathrm{TURNOVER\_RATIO} < 0.71 \land\\[-3pt]
\mathrm{CF\_TO\_ASSETS\_ewm\_mean} \le 0.02)\\[-3pt]
\lor\\[-3pt]
(\mathrm{AGE} \ge 9.00 \land \mathrm{CF\_TO\_ASSETS} \le 0.01)\\[1pt]
\end{gathered}$
\\ \hline

\end{tabularx}
\vspace{-0.5cm}
\end{table}

\noindent \textbf{Rule 1 –} This rule identifies firms beyond the early stages of their life cycle that give clear signs of financial fragility. In particular, it highlights negative retained earnings or extremely low cash-flow generation relative to assets, both of which indicate limited capacity to generate internal financial resources. The age threshold is economically meaningful, as abnormal financial ratios are relatively common among very young firms but become stronger distress signals for more mature firms.

\noindent \textbf{Rule 2 –} This rule captures alternative configurations of vulnerability. The first involves large and mature firms operating in sectors with high default rates, suggesting that sector-level distress may increase risk even for established firms. The other conditions identify severe liquidity shortages or very weak cash-flow generation. Overall, the rule indicates that default risk may arise from either firm-level fragility or adverse sectoral conditions.

\noindent \textbf{Rule 3 –} When macroeconomic variables are excluded, the model relies entirely on firm-level financial indicators. The rule highlights three signals of financial weakness: negative retained earnings, extremely low cash-flow generation, and very high leverage. Together, these conditions describe firms with fragile financial structures and strong dependence on external financing.

\noindent \textbf{Rule 4 –} This rule introduces the dynamic dimension of financial risk through historical information on firms' financial conditions. It identifies firms characterized by persistently negative retained earnings or extremely low cash-flow generation. These patterns indicate prolonged financial fragility rather than temporary performance fluctuations.

\noindent \textbf{Rule 5 –} This rule combines indicators of operational inefficiency with persistent liquidity weakness. Firms with low asset turnover and historically weak cash-flow generation appear particularly vulnerable, as poor operational performance translates into sustained financial fragility.\\

Overall, the extracted rules reveal consistent patterns associated with high default risk across the different dataset configurations. Despite being derived from datasets with different informational structures and predictive performance, the rules consistently highlight four key dimensions of financial vulnerability: weak internal liquidity generation, erosion of internal capital, high leverage, and poor operational performance. In addition, age thresholds indicate that such signals become particularly informative once firms move beyond the early stages of their life cycle. When contextual and historical variables are available, the rules also capture the role of sector distress and the persistence of financial weakness over time. These patterns suggest that the model identifies economically meaningful signals of financial distress rather than arbitrary statistical correlations.

\section{Conclusions}
\label{sec:conclusion}
This paper evaluated SME default prediction by combining ML models with XAI techniques in a high-dimensional, imbalanced financial environment. Using a large panel of Italian SME, the analysis demonstrates that XGBoost achieves the highest out-of-sample performance, outperforming traditional benchmarks. The results confirm that integrating macroeconomic and sectoral variables significantly improves predictive power over models relying solely on firm-level financial indicators.

Beyond performance, the study addresses the ``black box'' nature of ML in credit risk. The comparison of rule extraction methods shows that DEXiRE-EVO improves both predictive fidelity and alignment with contextual feature relevance compared to the original DEXiRE framework. By incorporating CIU-based relevance into its multi-objective evolutionary design, DEXiRE-EVO extracts rule sets that accurately replicate the model's logic while remaining economically interpretable.
The extracted rules reveal that default risk consistently stems from weak liquidity, capital erosion, excessive leverage, and poor operational performance. When contextual variables are included, the rules also capture sectoral distress and persistent instability, suggesting the model identifies coherent economic mechanisms.
Limitations and Future Research include: (i) Temporal Dynamics: Current historical modeling is preliminary; future research should explicitly model time-series evolution to capture richer financial trajectories. (ii) Private Credit Application: The framework is highly relevant to private credit markets, where lenders lack ``soft'' information and access to credit registries, making automated, interpretable tools essential. (iii) Survival Analysis: Future work could shift from fixed-horizon predictions to estimating the precise timing of default events to enhance early warning systems.

\begin{credits}
\subsubsection{\ackname}
This work was partially supported by the Validate-H Project (PInter 14-2025), the Spanish Ministry of Economy and Competitiveness (PID2023-147409NB-C22) funded by MCIN/AEI/10.13039/501100011033.
\end{credits}

%
%

\bibliographystyle{splncs04}
\bibliography{references}

\end{document}